\documentclass[manuscript,screen]{acmart}
\usepackage{multirow}
\usepackage{amsmath}

\usepackage{colortbl}
\usepackage{enumitem}
\usepackage{subfigure}
\AtBeginDocument{%
  \providecommand\BibTeX{{%
    \normalfont B\kern-0.5em{\scshape i\kern-0.25em b}\kern-0.8em\TeX}}}

\setcopyright{acmlicensed}
\copyrightyear{2018}
\acmYear{2018}
\acmDOI{XXXXXXX.XXXXXXX}

\acmConference[Conference acronym 'XX]{Make sure to enter the correct
  conference title from your rights confirmation emai}{June 03--05,
  2018}{Woodstock, NY}
\acmISBN{978-1-4503-XXXX-X/18/06}

\begin{document}

\title[Hierarchical Action Recognition]{Hierarchical Action Recognition: A Contrastive Video-Language Approach with Hierarchical Interactions}

\author{Rui Zhang}
\email{ruizhang@zhejianglab.com}
\orcid{0000-0001-9126-9790}
\affiliation{%
  \institution{Research Center for Data Hub and Security, Zhejiang Lab}
  \city{Hangzhou}
  \state{Zhejiang}
  \country{China}
  \postcode{311100}
}

\author{Shuailong Li}
\email{lsl@zhejianglab.com}
\affiliation{%
  \institution{Research Centre for Frontier Fundamental Studies, Zhejiang Lab}
  \city{Hangzhou}
  \state{Zhejiang}
  \country{China}
  \postcode{311100}
}

\author{Junxiao Xue}
\affiliation{%
  \institution{Research Center of Space based Computing System, Zhejiang Lab}
  \city{Hangzhou}
  \state{Zhejiang}
  \country{China}
  \postcode{311100}}
\email{xuejx@zhejianglab.com}

\author{Feng Lin}
\orcid{0000-0002-1199-5870}
\affiliation{%
  \institution{Research Centre for Frontier Fundamental Studies, Zhejiang Lab}
  \city{Hangzhou}
  \state{Zhejiang}
  \country{China}
  \postcode{311100}
}
\email{asflin@zhejianglab.com}

\author{Qing Zhang}
\affiliation{%
 \institution{Research Centre for Frontier Fundamental Studies, Zhejiang Lab}
  \city{Hangzhou}
  \state{Zhejiang}
  \country{China}
  \postcode{311100}
 }
 \email{qing.zhang@zhejianglab.com}

\author{Xiao Ma}
\affiliation{%
  \institution{Research Centre for Frontier Fundamental Studies, Zhejiang Lab}
  \city{Hangzhou}
  \state{Zhejiang}
  \country{China}
  \postcode{311100}
  }
\email{mx@zhejianglab.com}

\author{Xiaoran Yan}
\orcid{0000-0003-3481-1832}
\authornote{Corresponding author \\\textit{Email address:} \url{yanxr@zhejianglab.com} (Xiaoran Yan)}
\affiliation{%
  \institution{Research Center for Data Hub and Security, Zhejiang Lab}
  \city{Hangzhou}
  \state{Zhejiang}
  \country{China}
  \postcode{311100}}
\email{yanxr@zhejianglab.com}




\begin{abstract}
  Video recognition remains an open challenge, requiring the identification of diverse content categories within videos. Mainstream approaches often perform flat classification, overlooking the intrinsic hierarchical structure relating categories. To address this, we formalize the novel task of hierarchical video recognition, and propose a video-language learning framework tailored for hierarchical recognition. Specifically, our framework encodes dependencies between hierarchical category levels, and applies a top-down constraint to filter recognition predictions. We further construct a new fine-grained dataset based on medical assessments for rehabilitation of stroke patients, serving as a challenging benchmark for hierarchical recognition. Through extensive experiments, we demonstrate the efficacy of our approach for hierarchical recognition, significantly outperforming conventional methods, especially for fine-grained subcategories. The proposed framework paves the way for hierarchical modeling in video understanding tasks, moving beyond flat categorization.
\end{abstract}



\received{20 February 2007}
\received[revised]{12 March 2009}
\received[accepted]{5 June 2009}

\maketitle

\section{Introduction}
\label{sec:intro}
Human action recognition serves as a pivotal task in visual learning, contributing significantly to the advancement of video understanding~\cite{li2023videochat,liu2023visual}. In contrast to conventional unimodal recognition methodologies~\cite{DBLP:journals/ijon/LiuMXYTPG22,DBLP:journals/ijcv/KongF22}, recent research on human action recognition has focused on leveraging large-scale vision-language models (VLMs) to frame it as a video-text matching problem~\cite{DBLP:journals/pami/SunKRBWL23}.
Some approaches~\cite{DBLP:journals/corr/abs-2109-08472,DBLP:journals/corr/abs-2212-03640,DBLP:conf/eccv/NiPCZMFXL22} perform multi-modal learning using CLIP-based architectures~\cite{radford2021learning} to transfer representations from the image to video domain and employ a contrastive training scheme on video-prompt pairs, which have achieved great success in video recognition tasks and also promoted the progress of human action recognition tasks~\cite{9782511}.
However, these models mainly address flat recognition/classification tasks and do not consider hierarchical or multi-level recognition~\cite{silla2011survey}, which is more common in natural language processing~\cite{huang2019hierarchical,zhang2020matching} but also relevant in computer vision~\cite{shao2020finegym}. Hierarchical action recognition enables more fine-grained categorical distinctions critical for many applications. Furthermore, accounting for dependencies between hierarchical label levels during modeling facilitates reasoning over the hierarchical structure.

\begin{figure}[t]
  \centering
   \includegraphics[width=0.90\linewidth]{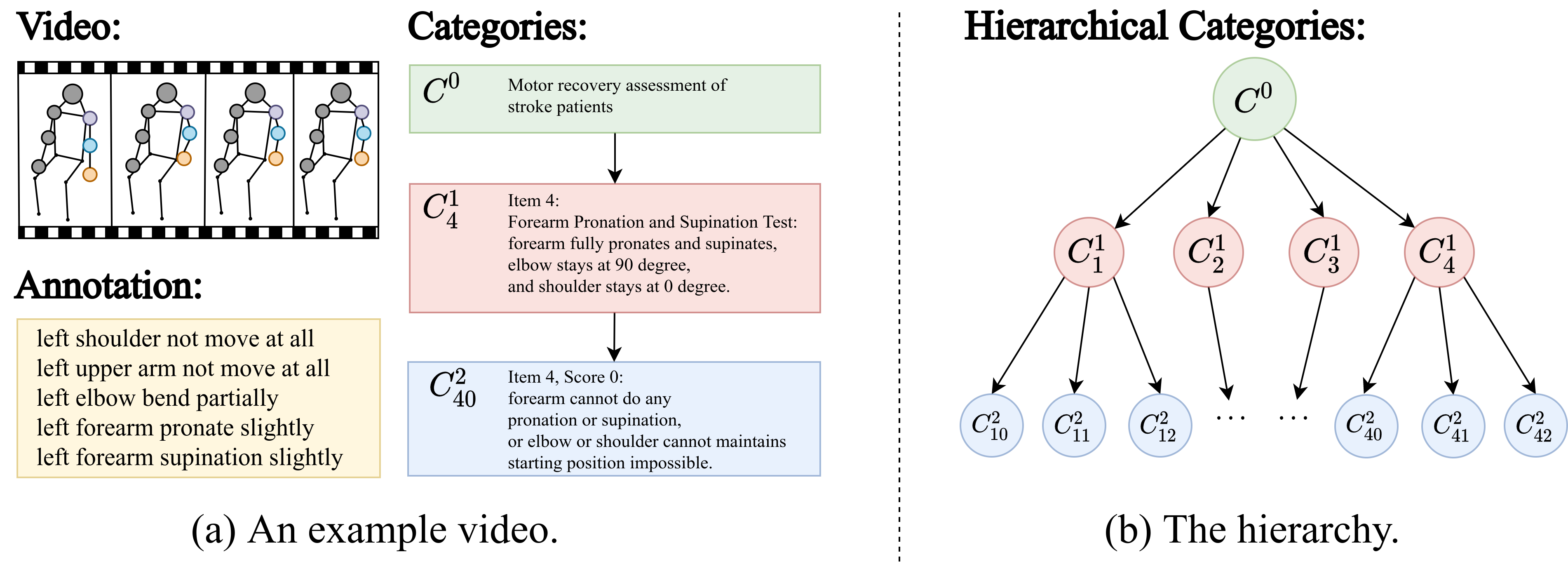}
   \caption{An illustrated example of a stroke patient undergoing motor recovery assessment in a hierarchical video recognition problem.}
   \label{fig:example}
\end{figure}

Numerous real-world applications entail the organization of objects in hierarchical structures, where classes are either subsumed into superclasses or refined into subclasses~\cite{bennett2009refined}. For instance, electronic books and movies are often associated with multiple hierarchical categories to enable more precise search and recommendation functions. 
A more prominent scenario for hierarchical recognition in the domain of computer vision is medical video-language understanding, which can facilitate clinical decision-making, augment patient engagement, and enhance patient health and well-being~\cite{gupta2022overview}. 
Figure~\ref{fig:example}a depicts an illustrative example of a stroke patient undergoing a motor recovery assessment. The video in the example comprises three levels of tags, namely the assessment summary, assessment items and assessment scores. Specifically, as shown in Figure~\ref{fig:example}b, the corresponding hierarchical categories are represented by a tree graph. The root node is the summary of the entire categories, which is considered as level 0, level 1 encompasses the detailed text descriptions of assessment items, and level 2 denotes the assessment scores of the relevant item. This hierarchical structure not only enables an elegant presentation of multi-layered data characteristics, but also provides a multi-dimensional perspective to address the recognition problem in a more structured manner. Although flat classification is sufficient in some basic cases, hierarchical recognition more closely resembles human conceptualization of knowledge and offers additional benefits like finer-grained distinctions and explicit parent-child relationships between categories across levels, which are particularly important in human-centric visual understanding. Therefore, advancing hierarchical action recognition capabilities poses an important direction for future research to improve video understanding tasks and align better with layered human cognition, and provide more accurate judgments and intelligent analysis in human-centric applications.

Since hierarchical action recognition is relatively nascent in the field of computer vision, we draws upon pertinent research in the domain of natural language processing and believe that this task is confronted  with two primary challenges: 1) the effective and rational utilization of hierarchical category structure information (e.g., \textit{Item 4, score 0: no pronation/supination} is a subclass of \textit{Item 4: pronation and supination assessment} as depicted in Figure~\ref{fig:example}b), and 2) the design of a model capable of capturing the dependencies between different levels in the hierarchy for interaction and fusion of hierarchical knowledge. While some existing research in the field of natural language processing has attempted to address these challenges~\cite{huang2019hierarchical,zhang2020matching}, the field of computer vision still lacks strategies to capture the associations between videos and hierarchical categories and consider the dependencies among different levels of the hierarchical structure. Furthermore, given the domain specificity of computer vision tasks compared to language, learning multi-modal knowledge with a hierarchical structure and executing hierarchical video recognition also pose significant difficulties in solving the problem of hierarchical action recognition.  Specifically, approaches need to account for the temporal nature of video data, handle both human (body)-centric visual and textual modalities, and model inter-level dependencies between categories - all while ensuring computational efficiency suitable for real-world usage. As research progresses, studying relevant methods from language domains could catalyze new techniques for modeling hierarchy, dependency, and multi-modality. Advances in this emerging subfield could enable fine-grained understanding across application areas such as medical diagnosis, education, industrial inspection, and more. Therefore, despite existing challenges, hierarchical action recognition warrants extensive investigation to unlock its potential.

To address these challenges and advance hierarchical video recognition, we propose a novel contrastive video-language framework, which is the first comprehensive attempt to employ contrastive learning for this task. To this end, we also construct a new dataset for the hierarchical video recognition task, based on the motor recovery assessment videos of stroke patients with fine-grained annotations of their body movements. 
Specifically, we first employ visual and text encoders to extract and represent the corresponding visual and linguistic knowledge, respectively. Then, we transfer the matching information between text descriptions of hierarchical categories and video segments in a level-by-level manner, and perform action recognition for each category level. In this process, we pay special attention to the exploitation and integration of fine-grained knowledge of body movements contained in each video clip and attempt to introduce fine-grained attention mechanisms to guide the learning of the model. The fine-grained matching also accounts for dependencies between hierarchy levels.  Finally, we devise an appropriate contrastive learning scheme to optimize our model and conduct experiments to validate the effectiveness of the method. 
Our approach opens up promising new research directions into contrastive learning for video domains. By learning joint video-text representations with hierarchical interactions, models can better capture the alignment of visual and linguistic knowledge at multiple semantic granularities, and provide more discriminative learning signals at each category level. 
Our main contributions are summarized as follows: 
\begin{itemize}
    \item To the best of our knowledge, this is the first comprehensive attempt to apply contrastive video-language learning for hierarchical action recognition problem which shows promising results.
    \item We devise a novel contrastive video-language learning framework with hierarchical interactions and emphasize on exploiting the fine-grained knowledge to guide the model learning. 
    \item We introduce a new challenging dataset with high-quality annotations and corresponding hierarchical labels to the community of hierarchical action  recognition for medical assessment. Specifically, our dataset focuses on motor function assessment in stroke rehabilitation patients.
    \item Extensive experiments are conducted to demonstrate the effectiveness of the proposed method on the dataset with hierarchically structured labels.
\end{itemize}

\section{Related Work}
\label{sec:formatting}
\subsection{Video Recognition}
Video recognition is a challenging task that requires large-scale and diverse datasets to train and evaluate models. Several datasets have been developed to enable video recognition research, such as UCF101~\cite{soomro2012ucf101}, ActivityNet~\cite{caba2015activitynet}, Kinetics~\cite{carreira2017quo}, Moments in Time~\cite{monfort2019moments} and so on, where some of them also provide annotations beyond category labels, e.g., like temporal locations~\cite{yeung2018every} or spatial-temporal bounding boxes~\cite{rodriguez2008action}. These large-scale video datasets have driven progress in developing video recognition models.
Over the last decade, video recognition models have evolved over time, starting with early attempts at adapting deep learning, then progressing to two-stream networks~\cite{simonyan2014two} that integrate optical flow, followed by the adoption of 3D convolutional kernels~\cite{6909619,Ng_2015_CVPR} to capture spatio-temporal features jointly, and finally leading to recent transformer-based architectures~\cite{DBLP:conf/iccv/Arnab0H0LS21,DBLP:conf/icml/BertasiusWT21,DBLP:conf/cvpr/LiuN0W00022,DBLP:conf/cvpr/YanXALZ0S22} that model long-range dependencies. These methods have achieved great success in video action recognition benchmark tasks.
However, most of these traditional works require a large amount of training data to perform a unimodal video classification task, and their high complexity and computational cost are still challenges in video recognition learning. Beyond standard recognition, more complex video understanding capabilities are still lacking, including hierarchical recognition, fine-grained analysis, human-object interactions, and instructional video comprehension. Developing models that can acquire these skills likely requires going beyond supervised categorization to incorporate relational knowledge mining, reasoning, and multimodal grounding. 

\subsection{Vision Language Models}
Vision-language models (VLMs) that jointly model visual and textual modalities have become a popular approach for video recognition. 
A notable example of this is CLIP~\cite{radford2021learning}, which has pioneered the large-scale pretraining of multi-modal transformers on 400 million image-text pairs. This approach learns alignments between visual and textual concepts, enabling transfer to a variety of downstream computer vision tasks for more precise and complete learning. 
As a result, the employment of VLMs for the purpose of video action recognition has increasingly emerged as a prevalent trend within the field. 
X-CLIP~\cite{DBLP:conf/eccv/NiPCZMFXL22} proposes a lightweight module that explicitly exchanges information across frames and also provides a video-specific prompting scheme to leverage video content information for generating discriminative textual prompts. And meanwhile, ViFi-CLIP~\cite{DBLP:journals/corr/abs-2212-03640} adapts the image-based CLIP model to the video domain through fine-tuning and shows that a simple fine-tuning of CLIP is sufficient to learn suitable video-specific inductive biases.
These methods leverage the rich generalized vision and language representations of CLIP, fusing them with additional components for temporal modeling to achieve robust performance on video recognition tasks.
Although these VLM methods have performed well on standard benchmarks, their objectives are often too broad and do not account for the particularities of fine-grained human action recognition in real-world settings. Challenges like occlusion, scale variation, subtle motion patterns, and complexity of human-object interactions are not sufficiently addressed. 
Although some works have begun tailoring VLMs to focus on salient spatial-temporal features for action recognition~\cite{wang2021end}, substantial room remains for improvement. 

\subsection{Vision Language Model in Action Recognition}
Action recognition is one of the most critical technologies in human-centered intelligent applications. From traditional works based on 2D and 3D convolutional neural networks~\cite{ji20123d,9290594,DBLP:conf/iclr/LiLWWQ21,DBLP:journals/pami/QiWSSHT23}, to recent transformer-based unimodal strategies~\cite{DBLP:conf/iccv/Arnab0H0LS21,DBLP:conf/icml/BertasiusWT21,DBLP:conf/cvpr/LiuN0W00022,DBLP:conf/cvpr/YanXALZ0S22}, the definition and solution of action recognition have continuously improved. This has culminated in action recognition emerging as a specialized research direction in computer vision.
In recent years, the rapid advancement of vision-language models has facilitated the emergence of many multi-modal approaches to action recognition. By considering the abundant semantic information contained in actions, these methods aim to address limitations of uni-modal models in ``few-shot'' and ``zero-shot'' transfer capabilities. 
ActionCLIP~\cite{DBLP:journals/corr/abs-2109-08472} is the first work that proposes a new paradigm, "pre-train, prompt and finetune", for video action recognition. A latest work, ASU~\cite{DBLP:journals/corr/abs-2303-09756} also draws on PaStaNet~\cite{DBLP:conf/cvpr/LiXLHXWFMCL20} and AKU~\cite{DBLP:conf/mm/MaWWLCLQ22} to further strengthen the role of semantic units (e.g., body parts, objects, and scenes) hidden behind action labels. 
These approaches have demonstrated the promise of joint video-text modeling on human action learning. However, these attempts have focused on flat classification, without exploring hierarchical video understanding.
Hierarchical recognition poses even greater difficulties due to the need for inter-level dependency modeling and fine-grained alignment of modalities. The only work in this context is HierVL~\cite{ashutosh2023hiervl}, which learns hierarchical video-language embeddings but targets scene graphs rather than human actions. It also does not fully consider category interactions and dependencies for prediction. Advancing hierarchical human action recognition requires specialized techniques to capture fine-grained semantics and motion dynamics, while relating them to hierarchical textual descriptors. This remains an open challenge despite the potential benefits for application areas such as healthcare, robotics, and education.

\subsection{Hierarchical Action Recognition}
In the field of computer vision, the concept of hierarchical structure was first explored by competitive sports video understanding tasks. To catalyze research, the community has curated benchmark datasets~\cite{shao2020finegym,xu2022finediving} and developed AI scoring systems~\cite{li2022dynamic,zhou2023hierarchical} for domains like gymnastics and diving. However, the objectives diverge from the hierarchical action recognition task proposed in this paper.
Specifically, competitive sports video understanding focuses on identifying fine-grained events throughout the video to judge perfection in executing complex moves and produce numerical scores. This is exemplified in springboard diving, where the diving system would evaluate the execution of an action sequence (e.g. 401B, 5331D, 5152B, 305B) to output a final score. In contrast, hierarchical video recognition aims to categorize the entire video content in a hierarchical taxonomy, recursively refining categorical tags from coarse to fine grains. For instance, in a medical assessment, the model would first identify the body part being evaluated, and subsequently determine the assessment score based on motion quality in the video.
Therefore, we propose the novel task of hierarchical human action recognition in real-world videos, and contribute solutions tailored to its unique challenges. These include modeling inter-level dependencies, fusing motion and language cues, and capturing fine-grained semantics. Advancing this frontier can potentially benefit healthcare applications like automated diagnostics, along with areas like human-robot interaction, surveillance, and multimedia retrieval.

Overall, hierarchical video understanding remains an open, exciting area at the intersection of multiple subfields. Developing architectures that connect granular motion patterns to hierarchical text descriptors could enable more human-like comprehension. Beyond precise categorization, future systems may acquire capabilities like visual reasoning, motion assessment, and causal understanding of human actions and interactions. Tackling the specific challenges of hierarchical modeling is a promising step toward this grander goal.
\begin{table*}[t]
\centering
\caption{Description of hierarchical FMA categories.}
\resizebox{0.99\linewidth}{!}{%
\begin{tabular}{|l|l|l|}
\hline
\multicolumn{3}{|c|}{\textbf{Fugl-Meyer Assessment}}                                                                                                                                                                                                                   \\ \hline
\multirow{16}{*}{\textbf{Items}} & \multicolumn{2}{l|}{\textbf{\begin{tabular}[c]{@{}l@{}}I1: Shoulder Flexion to 180° Test:\\       shoulder flexes from 90 degree to 180 degree, elbow stays at 0 degree, and forearm does not pronate or supinate.\end{tabular}}}   \\ \cline{2-3} 
                                 & \multirow{3}{*}{\textbf{Scores}}                                      & S0: shoulder abduction or elbow flexion happens immediately when shoulder flexion starts.                                                                   \\ \cline{3-3} 
                                 &                                                                       & S1: shoulder abduction or elbow flexion happens during movement.                                                                                            \\ \cline{3-3} 
                                 &                                                                       & S2: shoulder can flex to 180 degree, no shoulder abduction or elbow flexion.                                                                                \\ \cline{2-3} 
                                 & \multicolumn{2}{l|}{\textbf{\begin{tabular}[c]{@{}l@{}}I2: Shoulder Flexion to 90 Degree Test:\\      shoulder flexes from 0 degree to 90 degree, elbow stays at 0 degree, and forearm does not pronate or supinate.\end{tabular}}} \\ \cline{2-3} 
                                 & \multirow{3}{*}{\textbf{Scores}}                                      & S0: shoulder abduction or elbow flexion happens immediately when shoulder flexion starts.                                                                   \\ \cline{3-3} 
                                 &                                                                       & S1: shoulder abduction or elbow flexion happens during movement.                                                                                            \\ \cline{3-3} 
                                 &                                                                       & S2: shoulder can flexes to 90 degree, no shoulder abduction or elbow flexion.                                                                               \\ \cline{2-3} 
                                 & \multicolumn{2}{l|}{\textbf{\begin{tabular}[c]{@{}l@{}}I3: Shoulder Abduction to 90 Degree Test:\\      shoulder spreads out from 0 degree to 90 degree, elbow stays at 0 degree, and foream keeps neutral.\end{tabular}}}          \\ \cline{2-3} 
                                 & \multirow{3}{*}{\textbf{Scores}}                                      & S0: shoulder supination or elbow flexion happens immediately when shoulder abduction starts.                                                                \\ \cline{3-3} 
                                 &                                                                       & S1: shoulder supination or elbow flexion happens during movement.                                                                                           \\ \cline{3-3} 
                                 &                                                                       & S2: shoulder spread out 90 to degree, and maintains extension and pronation.                                                                                \\ \cline{2-3} 
                                 & \multicolumn{2}{l|}{\textbf{\begin{tabular}[c]{@{}l@{}}I4: Forearm Pronation and Supination Test:\\      forearm fully pronates and supinates, elbow stays at 90 degree, and shoulder stays at 0 degree.\end{tabular}}}             \\ \cline{2-3} 
                                 & \multirow{3}{*}{\textbf{Scores}}                                      & S0: forearm cannot do any pronation or supination, or elbow or shoulder cannot maintains starting position impossible.                                      \\ \cline{3-3} 
                                 &                                                                       & S1: forearm can do limited pronation and supination, and elbow and shoulder maintains starting position.                                                    \\ \cline{3-3} 
                                 &                                                                       & S2: forearm can pronates and supinates fully, and elbow and shoulder maintains starting position.                                                           \\ \hline
\end{tabular}
}
\label{tab:fma_full}
\end{table*}
\begin{table}[t]
\centering
\caption{The statistics of FMA dataset.}
\begin{tabular}{|c|c|c|c||c|c|}
\hline
\multicolumn{4}{|c||}{\textbf{Fugl-Meyer Assessment}}                                             & \multicolumn{2}{c|}{\textbf{Fully-supervised Learning}}                                                                                                 \\ \hline
\textbf{Item no.}            & \textbf{Score no.} & \textbf{\# videos} & \textbf{\# annotations} & training set: val set: test set                                                                              & 753: 252: 252                              \\ \hline
\multirow{3}{*}{\textbf{I1}} & \textbf{S0}        & 110                & 110                     & \multicolumn{2}{c|}{\multirow{2}{*}{\begin{tabular}[c]{@{}c@{}}\textbf{Few-shot Learning}\\ ($K$ training samples of each $C^2$-level category)\end{tabular}}} \\
                             & \textbf{S1}        & 114                & 114                     & \multicolumn{2}{c|}{}                                                                                                                                   \\ \cline{5-6} 
                             & \textbf{S2}        & 112                & 112                     & \multirow{2}{*}{\begin{tabular}[c]{@{}c@{}}training set: test set\\ ($K=2$)\end{tabular}}                 & \multirow{2}{*}{24: 1233}                \\ \cline{1-4}
\multirow{3}{*}{\textbf{I2}} & \textbf{S0}        & 109                & 109                     &                                                                                                              &                                          \\ \cline{5-6} 
                             & \textbf{S1}        & 112                & 112                     & \multirow{2}{*}{\begin{tabular}[c]{@{}c@{}}training set: test set \\  ($K=4$)\end{tabular}}                & \multirow{2}{*}{48: 1209}                \\
                             & \textbf{S2}        & 112                & 112                     &                                                                                                              &                                          \\ \hline
\multirow{3}{*}{\textbf{I3}} & \textbf{S0}        & 85                 & 85                      & \multirow{2}{*}{\begin{tabular}[c]{@{}c@{}}training set: test set\\  ($K=8$)\end{tabular}}                 & \multirow{2}{*}{96: 1161}                \\
                             & \textbf{S1}        & 84                 & 84                      &                                                                                                              &                                          \\ \cline{5-6} 
                             & \textbf{S2}        & 85                 & 85                      & \multirow{2}{*}{\begin{tabular}[c]{@{}c@{}}training set: test set\\  ($K=16$)\end{tabular}}                & \multirow{2}{*}{192: 1065}               \\ \cline{1-4}
\multirow{3}{*}{\textbf{I4}} & \textbf{S0}        & 112                & 112                     &                                                                                                              &                                          \\ \cline{5-6} 
                             & \textbf{S1}        & 111                & 111                     & \multicolumn{2}{c|}{\textbf{Zero-shot Learning}}                                                                                                        \\ \cline{5-6} 
                             & \textbf{S2}        & 111                & 111                     & test set                                                                                                     & 1257                                     \\ \hline
\end{tabular}
\label{tab:FMA_stat}
\end{table}
\section{Preliminary}
\subsection{Dataset Preparation}
\subsubsection{Category Construction.}
Medical assessment is the most ubiquitous application of hierarchical
video recognition. Hence, we focus our study on the Fugl-Meyer Assessment (FMA) - a widely adopted performance-based scale for quantifying motor impairment in post-stroke patients. 
The FMA comprises a standardized set of tasks used to evaluate disease severity and track recovery progress. It assesses multiple body parts on a 3-point scale of motor functioning. We restructure the FMA into a hierarchical tree, providing a systematic taxonomy of assessment items and scores.
Specifically, we carefully select four representative upper limb assessment categories, each with three ordered assessment scores indicating the evaluation degree. A score of 0 denotes an inability to perform the task, 1 indicates partial completion, and 2 represents full completion.
Table~\ref{tab:fma_full} provides an overview of the FMA hierarchy, comprising the four assessment items and three scoring levels per item. Table~\ref{tab:FMA_stat} details dataset statistics and train-validation-test splits used in experiments.

This hierarchical restructuring of a standard medical assessment provides a promising foundation for validating hierarchical video recognition methods. The real-world applicability helps demonstrate potential healthcare impacts.
Our work aims to provide an initial hierarchical structure to motivate further efforts in collecting multi-level clinical video data and developing real-world healthcare applications.

\subsubsection{Data Construction.}
We conduct a comprehensive video collection study with 28 adult volunteers aimed at capturing the selected FMA assessment categories. For each category, subjects perform the specified tasks while being recorded from multiple camera angles to provide diverse perspectives. Videos are examined for quality control and exclusion criteria of occlusion, motion blur, and inadequate framing, resulting in a final curated set of 1,257 high-quality video examples covering the full designed FMA hierarchy. 

The dataset comprises four assessment item categories mainly evaluating upper extremities. Under each item, there are three ordered assessment scores indicating the degree of successful task completion on a scale of 0 to 2. This provides a standardized taxonomy for training and benchmarking hierarchical models.
Furthermore, we manually annotate each video with fine-grained body part state descriptions. For the upper limb analysis, eight key parts are annotated - left and right shoulder, elbow, and forearm. Verbs describe the motions while adverbs indicate completeness. These atomic annotations are composed into full natural language sentences providing rich details of the motions and actions demonstrated.

This unique dataset combines standardized assessments, videos, and detailed natural language annotations. These rich multi-modal signals help models learn salient patterns for fine-grained hierarchical medical video analysis.

\subsection{Problem Formulation}
In this work, our central objective is hierarchical action recognition - automatically assigning videos to multi-level action categories based on a predefined taxonomy. Specifically, we aim to label videos from our constructed dataset with the relevant assessment items and scores through supervised model training.

We formalize each video $v_i$ in the set $V = \{v_1, v_2, \ldots, v_n\}$ as a sequence of frames $v_i = \{f_{v_i,1}, f_{v_i,2}, \ldots, f_{v_i,n_{v_i}}\}$. The annotation $a_i$ of video $v_i$ is an element of the natural language description set $A= \{a_1, a_2, \ldots, a_n\}$.
The target hierarchical taxonomy $\mathbb{C}=\{C^0, C^1, C^2\}$ contains three levels, where $C^0=\{c^0\}$ denotes the high-level assessment summary, $C^1=\{I1,I2,I3,I4\}$ represents the assessment items, and $C^2=\{I1S0, I1S1, I1S2,\ldots, I4S2\}$ represents the assessment scores. 
In practice,  $c^0$ denotes ``motor recovery assessment of stroke patients'', which is fixed, so we mainly focus on $C^1$ and $C^2$ for video recognition learning. In this paper, we also use the simplified notation $C^1_{*}$ and $C^2_{*}$ to denote the individual elements within sets $C^1$ and $C^2$ respectively (e.g., $C^1_{1}$ refers to the first assessment item category $I1$, and $C^2_{10}$ refers to $I1S0$ the first scoring level under $I1$ category).
Crucially, inter-level constraints exist between $C^1$ and $C^2$ - the sub-categories under any $C^1$ item can only be its corresponding score set, which enables the hierarchical relationships encoding.
Therefore, a set of $n$ video with expected hierarchical categories can be denoted as:
\begin{align}
    \mathcal{X} =\{(v_1, a_1, c^1_1, c^2_1),  \dots, (v_n, a_n, c^1_n, c^2_n)\}
\end{align}
where $ c^1_i \in C^1$ and $ c^2_i \in C^2$ are the expected categories from $\mathbb{C}$ assigned to the videos.
The target is to match each video $v$ with the corresponding hierarchical category structure $\mathbb{C}$. This is achieved by learning a multi-modal recognition model $\Phi$, which is used to predict the hierarchical action labels by capturing the semantics and structure,
\begin{equation}
    \label{eq:problem}
    \Phi(v, a, \mathbb{C}, \Theta) \rightarrow (c^1, c^2)
\end{equation}
where $\Theta$ denotes model parameters.

\section{Proposed Method}
\begin{figure*}[t]
  \centering
   \includegraphics[width=0.95\linewidth]{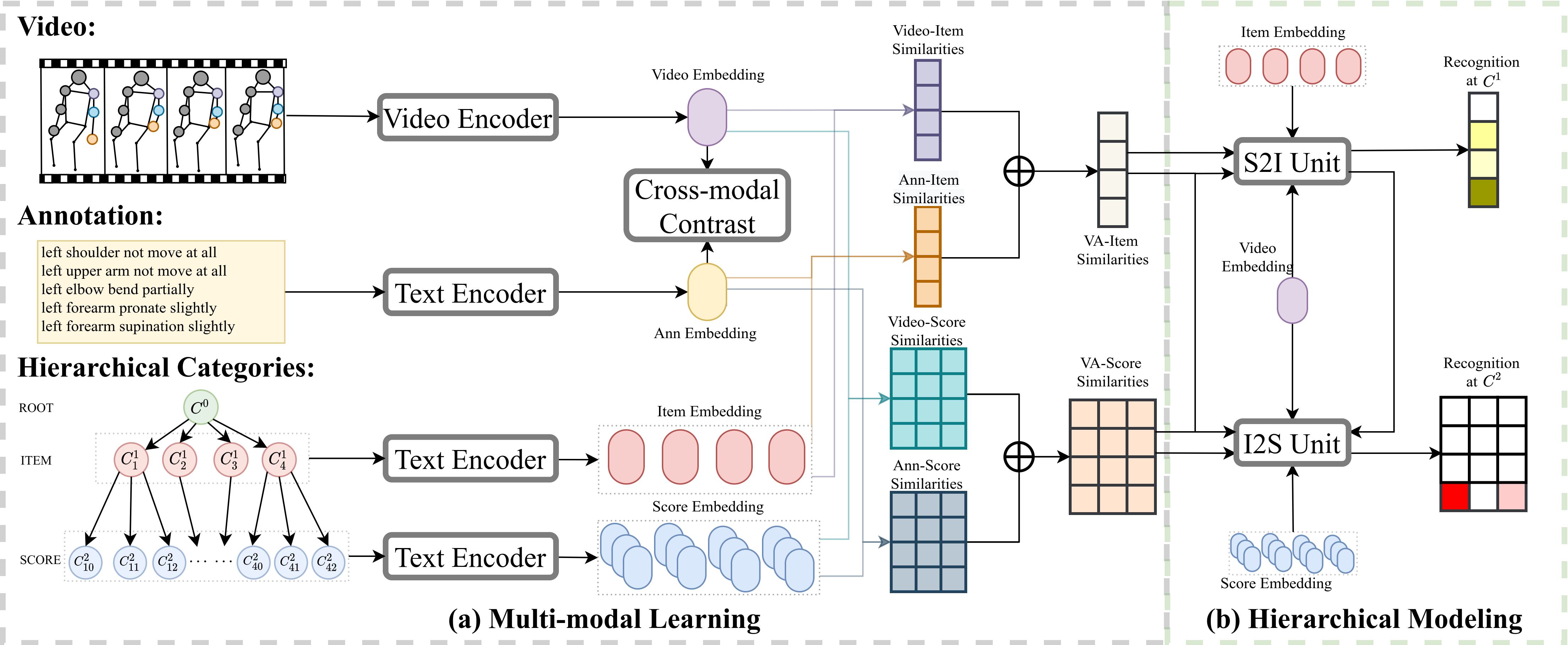}
   \caption{An overview of our proposed framework. We input videos, annotations, and descriptions of hierarchical categories and output the recognition results at each category level by two modules: multi-modal learning and hierarchical modeling , which are designed for facilitating knowledge learning at each level and enabling interaction among different levels of knowledge respectively. 
   }
   \label{fig:framework}
\end{figure*}

\begin{figure*}[t]
  \centering
   \includegraphics[width=0.95\linewidth]{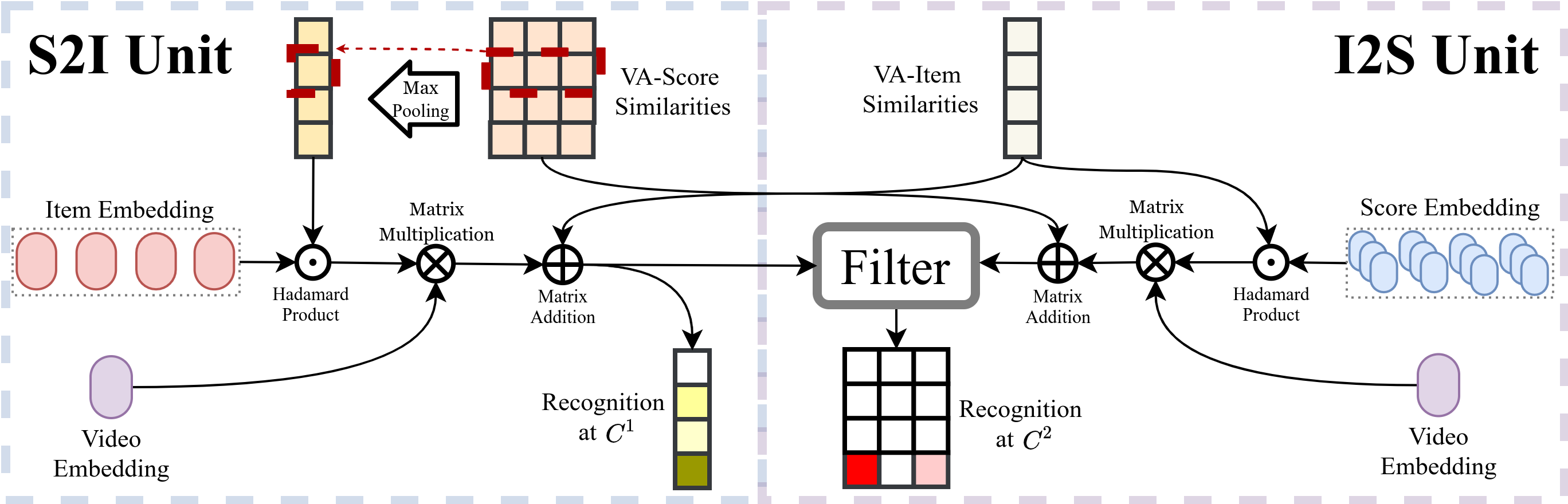}
   \caption{An illustration of S2I unit (left) and I2S unit (right).}
   \label{fig:units}
\end{figure*}

In this section, we introduce our contrastive video-language learning approach for hierarchical video recognition, which facilitates the hierarchical Fugl-Meyer assessment of videos by enabling knowledge interaction across a hierarchy of categories. We employ the CLIP model for general multi-modal encoding, leading to the development of a new model, which we refer to as H-CLIP. 
As depicted in Figure~\ref{fig:framework}, H-CLIP primarily comprises two components: multi-modal learning (Sec.~\ref{sec:mml}) and hierarchical modeling (Sec.~\ref{sec:hm}). The multi-modal learning component encodes and amalgamates multi-modal knowledge (comprising videos, video annotations, and textual descriptions of hierarchical categories) to derive their corresponding representations and compute the similarity scores among them. Hierarchical modeling is designed to capture the dependencies among different levels and the association between videos and each category within the hierarchical structure, thereby facilitating knowledge interactions. 
Besides, we believe that the results of upper-level categories exert a significant influence on lower-level predictions, and hence, we design a filter in the I2S unit to regulate hierarchical predictions in a top-down manner as shown in Figure~\ref{fig:units}.

\subsection{Multi-modal Learning}
\label{sec:mml}
The multi-modal learning module is divided into two key components, namely knowledge encoding and knowledge fusion. Knowledge encoding leverages the pre-trained image and text encoders from the CLIP model to embed the video frames and annotation sentences into a shared semantic space. This aligned embedding allows matching visual and linguistic concepts.
The subsequent knowledge fusion phase then aggregates these encoded representations to learn and integrate joint video-language knowledge independently at each level of the category hierarchy. This enables level-specific concept modeling and matching, handling inter-level dependencies.

\subsubsection{Knowledge Encoding.} 
To acquire multi-modal knowledge, we utilize the image encoder ($f_{img}$) and text encoder ($f_{txt}$) from the CLIP model, generating corresponding representations. For video data, we augment the image encoder with a temporal encoder ($f_{tem}$) to process temporal information within each video. This temporal encoder is a mean pooling function. Consequently, we obtain the following normalized embeddings:
video embedding $\mathbf{H}_{V}=f_{tem}\left(f_{img}\left(V\right)\right) \in \mathbb{R}^{n \times d}$, annotation embedding $\mathbf{H}_{A}=f_{txt}\left(A\right) \in \mathbb{R}^{n \times d}$, $C^1$-level category embedding $\mathbf{H}_{C^1}=f_{txt}\left(C^1\right) \in \mathbb{R}^{|C^1| \times d}$, and $C^2$-level category embedding $\mathbf{H}_{C^2}=f_{txt}\left(C^2\right) \in \mathbb{R}^{|C^1| \times 3 \times d}$, where $n$ is the number of samples, $d$ is the dimension of the knowledge representation, and $\mathbf{H}_{C^2}$ is a three-dimensional matrix since assessment scores depend on upper-level assessment items. 
To optimize the multi-modal embedding space, we calculate similarities between video and annotation representations and employ a symmetric InfoNCE loss~\cite{oord2018representation} on the similarity matrix for cross-modal contrast, which can be formulated as follows:
\begin{equation}
\label{eq:cross-modal_sim}
    \mathbf{S}_{V-A} = \mathbf{H}_{V} \cdot \mathbf{H}_{A}^\intercal \quad \in \quad \mathbb{R}^{n \times n}
\end{equation}
\begin{equation}
\label{eq:cross-modal_loss}
    \mathcal{L}_{cross}=-\frac{1}{n} \sum_{i=1}^n \log \frac{\exp \left(\mathbf{S}_{V-A}(i,i)\right)}{\sum_{j=1}^n \exp \left(\mathbf{S}_{V-A}(i,j)\right)}
\end{equation}

\subsubsection{Knowledge Fusion.}
In the context of hierarchical interactions, we first quantify the degree of resemblance between learning objects and categories at each level of the hierarchy independently. At the category level $C^*$, we compute the similarity matrix between the video representation $\mathbf{H}_V$ and the representation of categories $\mathbf{H}_{C^*}$:
\begin{equation}
    \mathbf{S}_{V-C^*} = \mathbf{H}_{V} \cdot \mathbf{H}_{C^*}^\intercal 
\end{equation}
As a result, $C^1$-level similarity matrix $\mathbf{S}_{V-C^1} \mathbb{R}^{n \times |C^1|}$ and $C^2$-level similarity matrix $\mathbf{S}_{V-C^2} \mathbb{R}^{n \times |C^1| \times 3}$ are generated.
Likewise, we calculate the similarity matrix between the annotation representation $\mathbf{H}_A$ and the category representation:
\begin{equation}
    \mathbf{S}_{A-C^*} = \mathbf{H}_{A} \cdot \mathbf{H}_{C^*}^\intercal 
\end{equation}
And the obtained similarity matrices are $\mathbf{S}_{A-C^1} \mathbb{R}^{n \times |C^1|}$ and $C^2$-level similarity matrix $\mathbf{S}_{A-C^2} \mathbb{R}^{n \times |C^1| \times 3}$.
We then average the similarity matrices to represent the assessment results of videos at the $C^*$ level:
\begin{equation}
    \mathbf{S}_{VA-C^*} = (\mathbf{S}_{V-C^*} + \mathbf{S}_{A-C^*}) / 2
\end{equation}
In the above calculations, $\mathbf{S}_{VA-C^1} \in \mathbb{R}^{n \times |C^1|}$ and $\mathbf{S}_{VA-C^2}\in \mathbb{R}^{n \times |C^1| \times 3}$ are similarity matrices determined independently at each corresponding level. These are going to serve as key components in the subsequent hierarchical interaction modeling.

\subsection{Hierarchical Modeling}
\label{sec:hm}
After obtaining independent similarity scores between videos and category levels, we conduct hierarchical interactions to model the structured relationships within the taxonomy, accounting for inter-level dependencies. This involves two key components operating bidirectionally across the hierarchy:
\begin{itemize}
\item S2I (score-to-item) units: Propagate relevance from fine-grained score levels to their corresponding item categories. This enables higher-order assessment items to accumulate evidence from the multiple scores under them.
\item I2S (item-to-score) units: Transfer top-down signals from assessment item categories to their associated score levels below. This allows higher-order context to guide lower-level matching.
\end{itemize}
As illustrated in Figure~\ref{fig:units}, these units connect the similarity scoring phases, enabling bidirectional propagation of relevance along the hierarchical edges. By explicit modeling of taxonomy constraints and inter-dependencies, the fused multi-modal knowledge becomes structured for coherent video-to-category assignment.

\subsubsection{S2I Unit.}
The S2I unit transfers fine-grained relevance from lower scoring levels to their associated higher-order assessment items and fuses it with the independent $C^1$-level matching information $\mathbf{S}_{VA-C^1}$ to obtain the final $C^1$-level recognition output. Specifically, we first apply max pooling on $\mathbf{S}_{VA-C^2}$ to reduce its dimensionality to the $C^1$ level, retaining and transferring only the strongest $C^2$-level matches to the $C^1$ calculation. This enables bottom-up interaction between hierarchical categories. We then integrate this compressed information with the $C^1$ category representations. The entire process can be described as:
\begin{equation}
\begin{split}
    \mathbf{MS}_{C^2} &= MaxPooling(\mathbf{S}_{VA-C^2}) \\
    \mathbf{NS}_{C^1} &= \mathbf{H}_{C^1} \odot \mathbf{MS}_{C^2}
\end{split}
\end{equation}
where $\odot$ is the Hadamard product between $\mathbf{H}_{C^1}$ and $\mathbf{MS}_{C^2}$ based on the $C^1$ dimension. 
To further connect videos and labels within this bottom-up interaction, we recompute their similarities:
\begin{equation}
\begin{split}
    \mathbf{S}_{C^1}^{'} &= \mathbf{NS}_{C^1} \cdot \mathbf{H}_{V} \quad  \in \quad \mathbb{R}^{n \times |C^1|}
\end{split}
\end{equation}
We fuse the independent $\mathbf{S}_{VA-C^1}$ and interactive $\mathbf{S}_{C^1}^{'}$ similarities via convex combination for the $C^1$-level prediction of the assessment items:
\begin{equation}
\begin{split}
    \hat{\mathbf{S}}_{C^1} = [\delta(\mathbf{S}_{VA-C^1}) + \delta(\mathbf{S}_{C^1}^{'})] / 2
\end{split}
\end{equation}
where $\delta$ applies the $softmax$ function along rows, normalizing similarities and ensuring values sum to one.

\subsubsection{I2S Unit.}
Along with the bottom-up interaction (S2I), we consider the top-down interaction (I2S) to allow assessment item $C^1$ predictions to influence lower-level $C^2$ scoring.
We map the $C^1$ similarities $\mathbf{S}{VA-C^1} \in \mathbb{R}^{n \times |C^1|}$ to the $C^2$ level by repeating each row three times along the $C^1$ dimension, yielding a mapped matrix $\mathbf{MS}{C^1} \in \mathbb{R}^{n \times |C^1| \times 3}$ that matches the shape of the $C^2$ tensor.
This is integrated with $C^2$-level embeddings $\mathbf{H}_{C^2}$ via element-wise product operation:
\begin{equation}
\begin{split}
    \mathbf{NS}_{C^2} &= \mathbf{H}_{C^2} \odot \mathbf{MS}_{C^1}
\end{split}
\end{equation}
Subsequently, video-oriented information matching is performed during hierarchical interaction to generate the $C^2$-level recognition result:
\begin{equation}
\begin{split}
    \mathbf{S}^{'}_{C^2} &= \mathbf{NS}_{C^2} \cdot \mathbf{H}_{V} \quad  \in \quad \mathbb{R}^{n \times |C^1| \times 3}
\end{split}
\end{equation}
As with $C^1$, we follow the same computation at the second category level of assessment scores ($C^2$) to fuse the independent and interactive terms for $C^2$, which is formulated as follows:
\begin{equation}
\begin{split}
    \hat{\mathbf{S}}_{C^2} &=  [\delta(\mathbf{S}_{VA-C^2}) + \delta(\mathbf{S}_{C^2}^{'})] / 2
\end{split}
\end{equation}
In light of the hierarchical constraints inherent between category levels, it is evident that lower-level categories exhibit significant affiliations with their corresponding upper-level categories. Consequently, we posit that it is not judicious to make independent evaluations for categories at each hierarchical level in the task of hierarchical recognition. Instead, the predictions made at the upper-level should exert a direct influence on the outcomes at the lower level.
To address this issue, we introduce an auxiliary filter, denoted as $\Omega$ within the I2S unit. This filter is designed to impose top-down constraints, i.e., retaining lower-level categories only for the maximal parent item. This restricts incoherent combinations violating the hierarchy, and the mathematical representation of this process is as follows:
\begin{equation}
    \hat{\mathbf{S}}_{C^2}^{'} = \Omega(\hat{\mathbf{S}}_{C^2}) \odot \hat{\mathbf{S}}_{C^2}
\end{equation}
Specifically, $\Omega$ filters $\hat{\mathbf{S}}_{C^2}$ based on the top $C^1$ assignment per video. Thereby, only compatible $C^2$ predictions are preserved after I2S. Each element $\Omega(\hat{\mathbf{S}}_{C^2}(i,j,k))$ is defined as follows:
\begin{equation}
    \Omega(\hat{\mathbf{S}}_{C^2}(i,j,k))= 
    \begin{cases}
        1 & \text{if } j = \underset{u}{\mathrm{argmax}}\, (\hat{\mathbf{S}}_{C^1} (i,u) )\\        
        0 & \text{otherwise}
    \end{cases}
\end{equation}
The primary objective of the filter is to retain only the assessment item that achieves the highest scores at the $C^1$ level. This approach ensures that the lower-level predictions are evaluated only when the upper-level recognition is accurate. Conversely, if the identification at the upper level is incorrect, the prediction at the lower level is immediately deemed to be erroneous.

In total, bidirectional I2S and S2I units allow hierarchical interactions between video-label matching rounds. The filtering mechanism further enforces coherence for taxonomy-structured recognition.

\subsubsection{Optimization.}
To optimize our proposed KG-CLIP, we employ the Kullback-Leibler (KL) divergence as the video-text contrastive loss at each category level. We construct ground-truth matrices $\mathbf{G}_{C^1}$ and $\mathbf{G}_{C^2}$, where positive pairs have a target similarity of 1, while negative pairs have a target of 0.
Based on this, we formulate the hierarchical loss as follows:
\begin{equation}
\begin{split}
    \mathcal{L}_{hier}^1 &= \mathcal{D}_{KL}(\delta(\hat{\mathbf{S}}_{C^1}) || \delta(\mathbf{G}_{C^1})) \\
    \mathcal{L}_{hier}^2 &= \mathcal{D}_{KL}(\delta(\hat{\mathbf{S}}_{C^2}^{'}) || \delta(\mathbf{G}_{C^2}))
\end{split}
\label{eq:kl}
\end{equation}
where $\delta$ applies softmax to normalize rows, and $\mathcal{D}_{KL}$ computes the KL divergence between prediction and target similarity distributions. Minimizing the KL loss pulls positive pairs closer to a similarity of 1, while pushing negative pairs to 0 in a probabilistic manner. Applying this hierarchically provides a structured objective.

The overall hierarchical recognition loss combines cross-modal and hierarchical terms:
\begin{equation}
    \mathcal{L} = \mathcal{L}_{cross} +  \lambda_1 \mathcal{L}_{hier}^1 +  \lambda_2\mathcal{L}_{hier}^2
\end{equation}
where $\lambda_{1,2} $ weight the influence of assessment item and scoring level losses. These can be tuned for optimal taxonomy-structured prediction.

\section{Experiments}

\subsection{Experimental Setup}
\subsubsection{Baselines.}
As our proposed method is the first work on hierarchical video recognition with a cross-modal contrastive framework, we compare against several recent CLIP-based multi-modal baselines and a different version of our approach:
\begin{itemize}
    \item X-CLIP\footnote{https://github.com/microsoft/VideoX/tree/master/X-CLIP}
    ~\cite{DBLP:conf/eccv/NiPCZMFXL22} transfers the CLIP model from the image domain to the video domain through a cross-frame attention mechanism to capture temporal information and inter-object relationships in the video.
    \item ViFi-CLIP\footnote{https://github.com/muzairkhattak/ViFi-CLIP}
    ~\cite{DBLP:journals/corr/abs-2212-03640} demonstrates a novel video-based CLIP model fine-tuning strategy that achieves good recognition results without any new modules or components.
    \item H-CLIP-t is a variant of H-CLIP, which uses a more complex Transformer as the temporal encoder and introduces more trainable parameters to optimize the modeling of the video temporal dimension.
\end{itemize}

It should be noted that the two baselines lack structured taxonomy-aware objectives and components. As a result, they perform flat classification on each category level independently, without any inter-level constraints.

\subsubsection{Implementation Details.}
\begin{table}[t]
\centering
\caption{The setup of (hyper-) parameters for H-CLIP experiments.}
\begin{tabular}{|l|r|}
\hline
\textbf{Seed}                        &   1024        \\ \hline
\textbf{Batch size}                  & 16, 32, 64      \\
\textbf{\# frames}                   & 8, 16       \\
\textbf{Input size of frames}        &   224        \\ \hline
\textbf{Backbone}                    & ViT-B/32, ViT-B/16 \\ \hline
\textbf{\# epoch}                    &     30      \\
\textbf{\# workers}                  &      8       \\
\textbf{Optimizer}                   &    Adam~\cite{DBLP:journals/corr/KingmaB14}       \\
\textbf{Learning rate (CLIP modules)} &     $1\mathrm{e}{-5}$   \\
\textbf{Learning rate (new modules)}    &     $1\mathrm{e}{-4}$     \\
\textbf{Warming up steps}            &      5       \\
\textbf{Weight decay}                &     0.2        \\ 
$\lambda_1$, $\lambda_2$               &     1.0, 1.0        \\ \hline
\textbf{RandAugment(N, M) for training~\cite{cubuk2020randaugment}}    &  N=2, M=9         \\ \hline
\end{tabular}
\label{tab:setup}
\end{table}
The text encoder and image encoder of all CLIP-based methods are initialized by the public CLIP checkpoints (ViT-B/32 with input patch sizes of 32 and ViT-B/16 with input patch sizes of 16). 
We utilize the AdamW optimizer~\cite{DBLP:journals/corr/KingmaB14} for model training, where the CLIP encoders are fine-tuned with an initial learning rate of $1\mathrm{e}{-5}$, 
and other new training modules use $1\mathrm{e}{-4}$. 
Learning rates are warmed up for 5 epochs then decayed to zero following a cosine schedule. 
We train our model for 30 epochs with a weight decay of 0.2.
For videos, we follow the same sampling strategy~\cite{DBLP:journals/pami/0002X00LTG19} and data augmentation~\cite{DBLP:journals/corr/abs-2109-08472} to process video frames into 224 × 224 spatial resolution. Experiments are conducted on a single NVIDIA A6000-48GB GPU. 

Table~\ref{tab:setup} shows the setup of (hyper-) parameters used in all H-CLIP experiments. Parameter settings are the same for all fully supervised learning and few-shot/zero-shot learning experiments.

\subsubsection{Evaluation Protocols.}
To evaluate the recognition performance of H-CLIP, we follow the previous video recognition work~\cite{DBLP:journals/corr/abs-2109-08472,DBLP:journals/corr/abs-2303-09756,DBLP:conf/eccv/LinGZGMWDQL22,DBLP:conf/eccv/NiPCZMFXL22,DBLP:journals/corr/abs-2212-03640}, using Top-1 accuracy (Top-1) and Top-3 accuracy (Top-3) as the recognition metrics for the category level of assessment items ($C^1$).
For the category level of assessment scores ($C^2$), we only report Top-1 accuracy since there are only three score child nodes under each assessment item.
The predictions of baselines on assessment items and assessment scores are completely independent, so their accuracy values are also calculated for 4 assessment items and 12 assessment scores respectively.

\subsection{Overall Performance}

\begin{table}[t]
\centering
\caption{Performance (\%) comparison on each category level ($C^1$ level and $C^2$ level) with state-of-the-art on FMA dataset. H-CLIP and its variant are the only methods that can perform simultaneous recognition on both levels, while the other methods need separate learning for each level.}
\begin{tabular}{|c|c|c|cc|c|}
\hline
\multirow{2}{*}{\textbf{Methods}} & \multirow{2}{*}{\textbf{Backbones}} & \multirow{2}{*}{\textbf{Frames}} & \multicolumn{2}{c|}{$C^1$ (Item)} & $C^2$ (Score) \\ \cline{4-6} 
                                  &                                     &                                     & Top-1                      & Top-3                      & Top-1                               \\ \hline\hline
\multirow{4}{*}{X-CLIP}           & \multirow{2}{*}{ViT-B/32}           & 8                                   & 84.92                      & {100.00}                     & 50.00                               \\
                                  &                                     & 16                                  & 88.49                      & {100.00}                     & 55.16                               \\ \cline{2-6} 
                                  & \multirow{2}{*}{ViT-B/16}           & 8                                   & 89.68                      & 99.60                      & 58.33                               \\
                                  &                                     & 16                                  & 88.49                      & 99.60                      & 58.33                               \\ \hline
\multirow{4}{*}{ViFi-CLIP}        & \multirow{2}{*}{ViT-B/32}           & 8                                   & 83.73                      & 99.21                      & 43.65                               \\
                                  &                                     & 16                                  & 82.54                      & 99.60                      & 41.67                               \\ \cline{2-6} 
                                  & \multirow{2}{*}{ViT-B/16}           & 8                                   & 86.11                      & 99.60                      & 48.41                               \\
                                  &                                     & 16                                  & 83.72                      & 99.60                      & 48.41                               \\ \hline\hline
\multirow{4}{*}{H-CLIP-t}        & \multirow{2}{*}{ViT-B/32}           & 8                                   & 96.83                      & {100.00}                     & 92.06                               \\
                                  &                                     & 16                                  & 97.22                      & {100.00}                     & 93.25                               \\ \cline{2-6} 
                                  & \multirow{2}{*}{ViT-B/16}           & 8                                   & {98.02}                      & {100.00}                     & 94.05                               \\
                                  &                                     & 16                                  & 97.22                      & {100.00}                     & 93.65                               \\ \hline
\rowcolor[HTML]{EFEFEF} 
\multicolumn{1}{|l|}{\cellcolor[HTML]{EFEFEF}}         &          & 8                                   & {98.02}                      & {100.00}                     & 92.46                               \\
\rowcolor[HTML]{EFEFEF} 
\multicolumn{1}{|l|}{\cellcolor[HTML]{EFEFEF}}                    &    \multirow{-2}{*}{ViT-B/32}                                   & 16                                  & {98.02}                      & {100.00}                     & 94.05                               \\ \cline{2-6} 
 \rowcolor[HTML]{EFEFEF} 
\multicolumn{1}{|l|}{\cellcolor[HTML]{EFEFEF}}               &           & 8                                   & {98.02}                      & {100.00}                     & 94.44                               \\
                   \rowcolor[HTML]{EFEFEF} 
\multicolumn{1}{|c|}{\multirow{-4}{*}{\cellcolor[HTML]{EFEFEF}H-CLIP}}   
 &       \multirow{-2}{*}{ViT-B/16}                               & 16                                  & {98.02}                      & {100.00}                     & {95.24 }                              \\ \hline
\end{tabular}
\label{tab:overall}
\end{table}
\subsubsection{Fully-supervised Recognition}
Table~\ref{tab:overall} shows the experimental results of our approach and two CLIP-based multi-modal baselines and a variant of our model. It is important to mention that the two baseline models, X-CLIP and ViFi-CLIP, are not designed to handle hierarchical video recognition tasks. Consequently, we partition the categories of levels $C^1$ and $C^2$ into two distinct tasks for these models to learn.
From the results, we can get several observations:
\begin{itemize}
    \item We evaluate the performance of our proposed H-CLIP and its variants against existing baselines on FMA dataset, which comprises two levels of categories ($C^1$ and $C^2$). Our method surpasses the baselines on both category levels, attesting to the effectiveness of our hierarchical recognition strategy. In particular, the $C^1$-level recognition achieves an enhancement of about 8\%, while the $C^2$-level recognition attains a remarkable enhancement of about 36\%.
    \item The baselines, which are not tailored for hierarchical video recognition tasks, perform category recognition at levels $C^1$ and $C^2$ independently. Although their recognition performance on the $C^1$-level category is acceptable, the performance on the more fine-grained and challenging $C^2$ level is considerably lower. Our proposed H-CLIP addresses this hierarchical recognition problem efficiently. By modeling hierarchical interactions, the predictions at the upper and lower levels mutually influence each other, resulting in a significant improvement in recognition performance. This improvement is especially evident at the more difficult lower-level categories.
    \item Despite H-CLIP-t, employing a Transformer to replace the mean pooling function for more careful encoding of temporal information within each video, its overall performance is slightly inferior to that of H-CLIP. We hypothesize that this is due to the limited number of training samples. Given the unique nature of the data in this study, the quantity of data is relatively limited, preventing the complex Transformer from being adequately trained and resulting in slightly poorer performance for this variant. Nevertheless, based on existing research~\cite{han2021transformer,khan2022transformers}, we are confident that H-CLIP-t holds great potential. We anticipate that with a sufficient number of training samples, it could deliver even better performance.
\end{itemize}

\subsubsection{Zero-/Few-shot Recognition.}
\begin{figure*}[tb]
  \centering
  \subfigure[Recognition at $C^1$ level.]{\label{fig:item}\includegraphics[width=110mm]{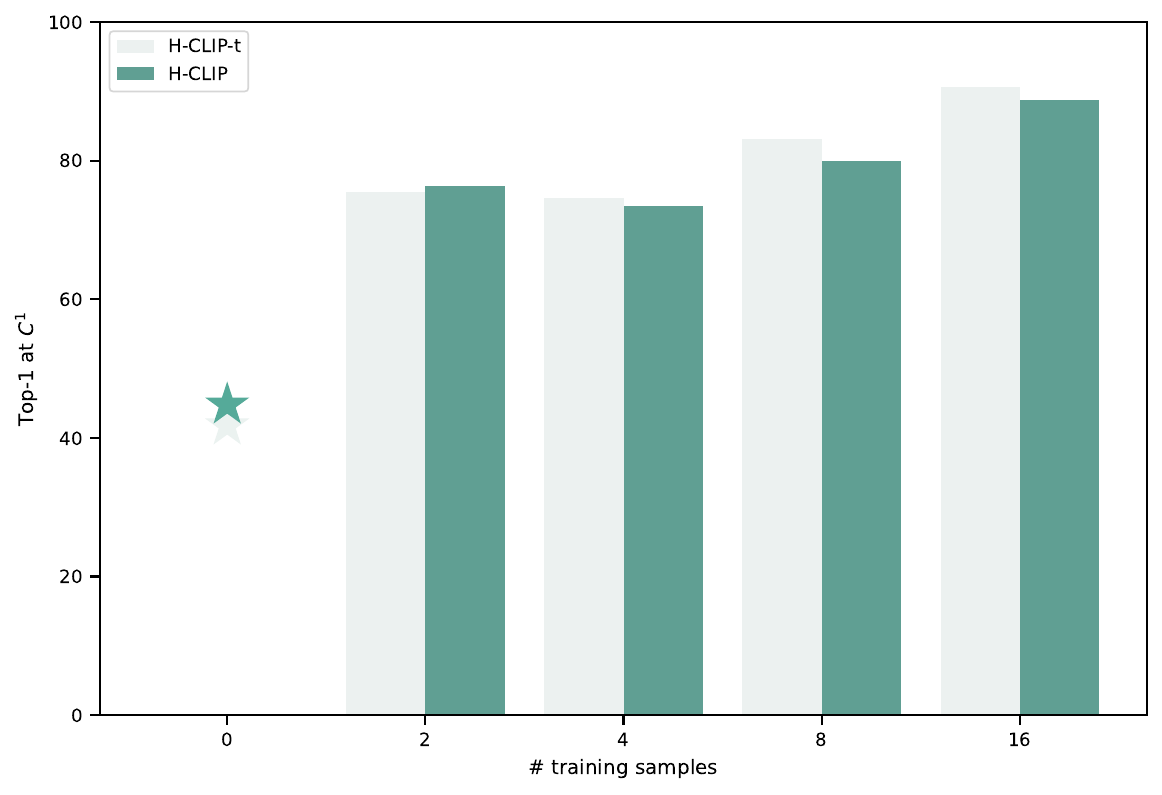}}
  \quad
  \subfigure[Recognition at $C^2$ level.]{\label{fig:score}\includegraphics[width=110mm]{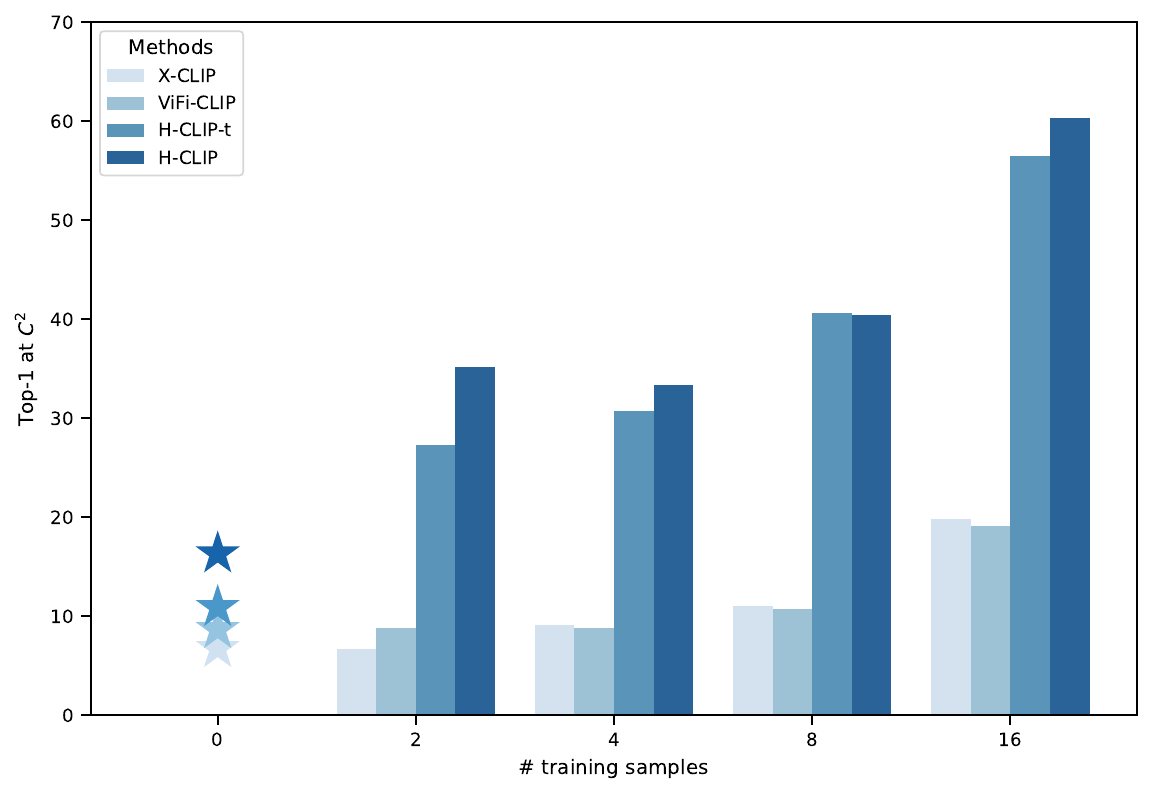}}
  \caption{Zero-/few-shot results on FMA dataset under 8 video frames with ViT-B/32 as the backbone. Two baselines are only capable of performing video recognition at the $C^2$ level, as they lack the ability to support hierarchical recognition.}
  \label{fig:zf}
\end{figure*}
We evaluate the zero/few-shot identification capabilities of H-CLIP. The experiment is conducted using 8 frames as the visual input of the model and ViT-B/32 as the backbone model. The number of training samples in this experiment is based on the $C^2$-level labels, so for the baselines that do not support hierarchical recognition, we only show their recognition results at the $C^2$ level. We present the performance of two baselines, the variant of H-CLIP and H-CLIP in Figure ~\ref{fig:zf}.
We observe that compared with the baselines, the zero/few-shot recognition performance of H-CLIP and its variant at the $C^2$ level is much higher, demonstrating the stronger transfer learning enabled by our hierarchical modeling. For the $C^1$ level, H-CLIP-t has comparable or better performance than H-CLIP. This is also due to the Transformer’s own modeling ability, which can capture potential connections that cannot be discovered by the mean pooling function after certain training.

The superior zero/few-shot identification of H-CLIP highlight the benefits of embedding structured relationships and inter-level transfer. By accounting for taxonomy semantics, the model achieves higher coherence and generalization.
These promising results motivate further research into hierarchical video-language architectures for low-resource recognition. As real-world video collection often lacks exhaustive labeling, such capabilities can facilitate adopting these models in practice. 
In summary, our experiments validate the usefulness of hierarchical modeling under data constraints through quantitative analysis and comparison. 

\subsection{Model Analysis}
\subsubsection{Ablation Study}
To further investigate the effectiveness of the proposed components in H-CLIP, we conduct ablation experiments under full supervision on the dataset. For experiments requiring videos, we sample 8 video frames as the visual input of the model. As shown in Table~\ref{tab:abl}, the complete H-CLIP achieves the best performance, validating the utility of each constituent.
Specifically,
\textbf{(1)} we first remove the cross-modal contrast (Equation~\ref{eq:cross-modal_loss}) to verify the importance of cross-modal learning in H-CLIP. We find that due to the powerful knowledge modeling capabilities of the pre-trained model, the generated data representation already contains rich connections between cross-modal knowledge, so the performance is still significant even without the cross-modal contrast.
\textbf{(2)} To verify the validity of our hierarchical learning strategy, we delete the hierarchical interactions (S2I unit and I2S unit) and conduct the experiments. We observe that compared with H-CLIP with complete components, its performance drops by about 1\% to 4\%, which proves that the hierarchical interaction mechanism we introduced has a significant positive effect on the recognition task.
For learning objects, we also examine the importance of different modal information. We perform the recognition task \textbf{(3)} using only visual knowledge and \textbf{(4)} using only annotated text. We see that the recognition accuracy in both experiments decreases to varying degrees, especially for the $C^2$ level categories. Compared with visual knowledge, annotations are more important, because unlike other data, the differences among videos of medical assessment are more subtle, and pure visual information is not enough to make accurate judgments. Therefore, we believe textual annotations are very important for video recognition of medical assessment.
\textbf{(5)} We also use cross-entropy loss to replace the KL loss in Equation~\ref{eq:kl}, and the results show about 10\% and 40\% decrease at the $C^1$ and $C^2$ levels respectively, which may be due to the superiority of KL divergence in quantifying distribution differences and discouraging over-confidence.

In total, our analysis verifies each component's efficacy and superiority over ablated variants. The vital roles of hierarchy and text are surfaced to facilitate precise medical video understanding. 
\begin{table}[t]
\centering
\caption{Ablation study on the effect of proposed components under 8 frames of each video, where KLL denotes KL loss and CEL denotes cross-entropy loss.}
\begin{tabular}{|lcc|}
\hline
\multicolumn{3}{|c|}{\textbf{Backbone: ViT-B/32}}                                                                                                                       \\ \hline
\multicolumn{1}{|c|}{{\textbf{Methods}}}         & \multicolumn{1}{c}{{Top-1 at $C^1$}} & {Top-1 at $C^2$} \\ \hline
\rowcolor[HTML]{EFEFEF} 
\multicolumn{1}{|l|}{\cellcolor[HTML]{EFEFEF}H-CLIP}   & 98.02                                                       & 92.46                                  \\
\multicolumn{1}{|l|}{-w/o cross-modal contrast}                   & 97.22                                                       & 91.67                                  \\
\multicolumn{1}{|l|}{-w/o hierarchical interactions}    & 94.44                                                       & 88.49                                  \\
\multicolumn{1}{|l|}{-w/o annotations}                       & 77.78                                                       & 38.89                                  \\
\multicolumn{1}{|l|}{-w/o videos}                  & 86.51                                                       & 82.54                                  \\
\multicolumn{1}{|l|}{-replaced KLL with CEL} & 83.33                                                       & 41.67                                  \\ \hline
\multicolumn{3}{|c|}{\textbf{Backbone: ViT-B/16}}                                                                                                                       \\ \hline
\rowcolor[HTML]{EFEFEF} 
\multicolumn{1}{|l|}{\cellcolor[HTML]{EFEFEF}H-CLIP}   & 98.02                                                       & 94.44                                  \\
\multicolumn{1}{|l|}{-w/o cross-modal contrast}                   & 97.62                                                       & 93.25                                  \\
\multicolumn{1}{|l|}{-w/o hierarchical interactions}    & 97.62                                                       & 92.06                                  \\
\multicolumn{1}{|l|}{-w/o annotations}                       & 90.87                                                       & 59.13                                  \\
\multicolumn{1}{|l|}{-w/o videos}                  & 84.13                                                       & 78.97                                  \\
\multicolumn{1}{|l|}{-replaced KLL with CEL} & 89.29                                                       & 47.22                                  \\ \hline
\end{tabular}
\label{tab:abl}
\end{table}

\begin{figure}
  \centering
  \subfigure[Recognition with different backbones.]{\label{fig:backbone}\includegraphics[width=110mm]{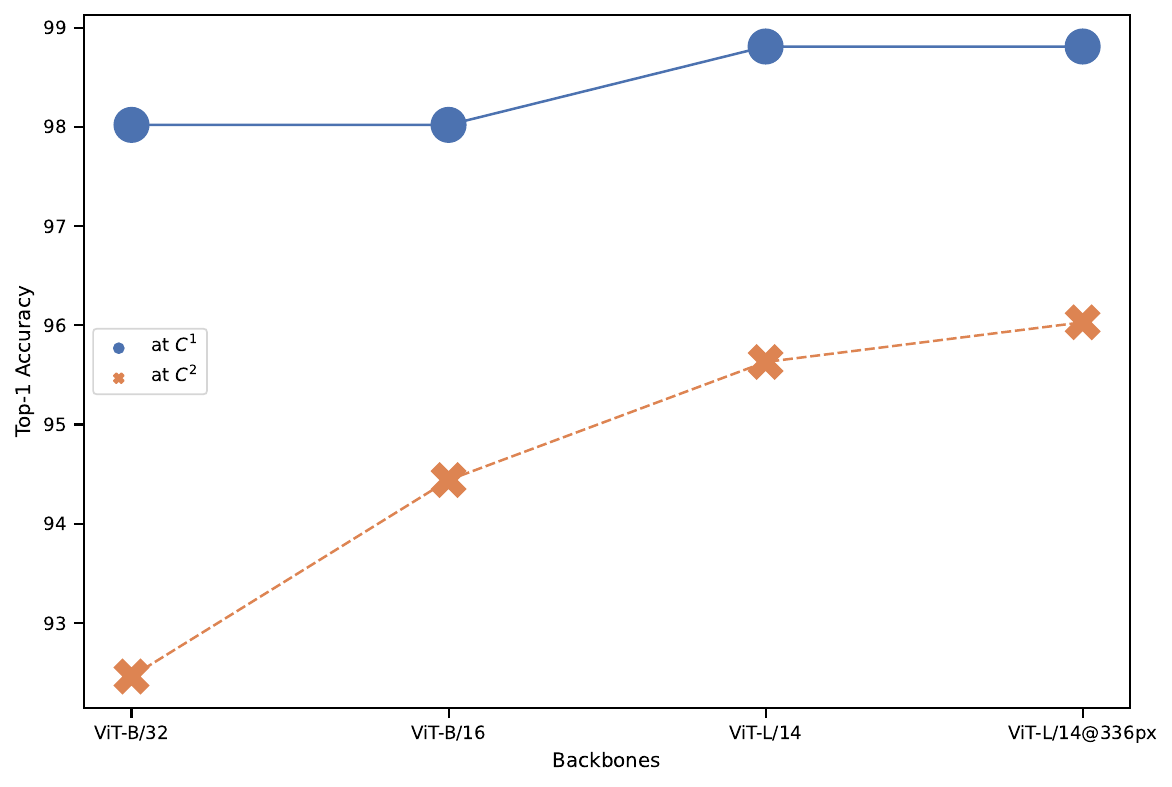}}
  \quad
  \subfigure[Recognition with different number of frames.]{\label{fig:frame}\includegraphics[width=110mm]{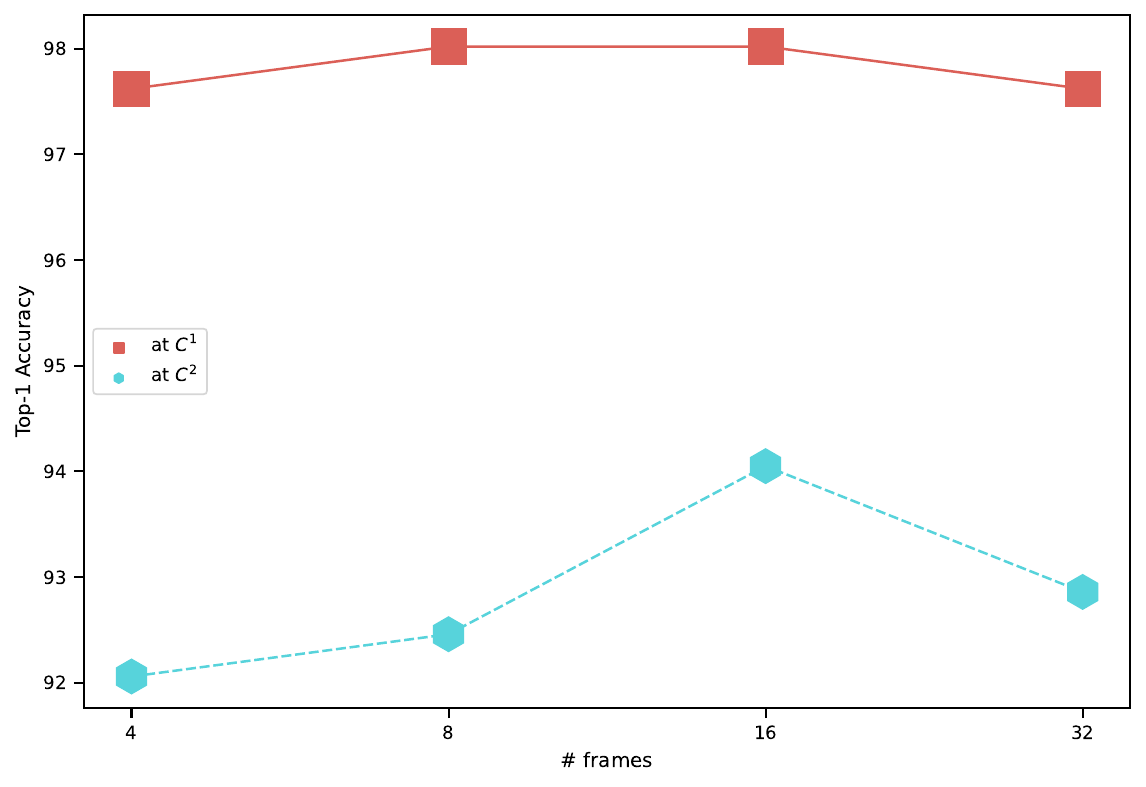}}
  \caption{Recognition results of different backbones under 8 video frames as the visual input and different number of frames under ViT-B/32 as the backbone model.}
  \label{fig:bf}
\end{figure}

\subsubsection{Analysis of Backbones and Number of Frames}
We conduct ablation studies to analyze the impact of backbone architecture and number of input frames on H-CLIP's performance.
Experiments are with varying backbone models including ViT-B/32, ViT-B/16, ViT-L/14 and ViT-L/14@336px under 8 input frames as shown in Figure~\ref{fig:backbone}, which reveal larger backbones consistently improve recognition, especially for fine-grained $C^2$ categories (~3\% gain).
Additional experiments sweeps input frames from 4 to 32 with a ViT-B/32 backbone. Figure~\ref{fig:frame}) demonstrates that the optimal frame number is data-dependent, without a clear universal trend. However, 8 and 16 frames often emerge as prudent choices for video recognition models.
In summary, H-CLIP demonstrates stable $C^1$ recognition across configurations. For the more challenging fine-grained $C^2$ task, performance is sensitive to backbone scale and input frames, but remains highly accurate (92\%+), highlighting the efficacy of our approach.

\subsubsection{Efficiency Study}
\begin{table}[t]
\centering
\caption{Model efficiency analysis under 8 frames of each video, with respect to FLOPs (G) and number of parameters (M).}
\begin{tabular}{|c|c|cc|}
\hline
\textbf{Methods}                                 & {\textbf{Backbones}} & {\textbf{FLOPs (G)}} & \textbf{\# params (M)} \\ \hline
                                                 & ViT-B/32                                & 106.96                                  & 196.6                  \\
\multirow{-2}{*}{X-CLIP}                         & ViT-B/16                                & 212.98                                  & 194.9                  \\ \hline
                                                 & ViT-B/32                                & 106.07                                  & 126                    \\
\multirow{-2}{*}{ViFi-CLIP}                      & ViT-B/16                                & 211.42                                  & 124.3                  \\ \hline
                                                 & ViT-B/32                                & 84.59                                   & 170.2                  \\
\multirow{-2}{*}{H-CLIP-t}                       & ViT-B/16                                & 184.59                                  & 168.6                  \\ \hline
\rowcolor[HTML]{EFEFEF} 
\cellcolor[HTML]{EFEFEF}                         & ViT-B/32                                & 84.44                                   & 151.2                  \\
\rowcolor[HTML]{EFEFEF} 
\multirow{-2}{*}{\cellcolor[HTML]{EFEFEF}H-CLIP} & ViT-B/16                                & 184.44                                  & 149.6                  \\ \hline
\end{tabular}
\label{tab:eff}
\end{table}
To evaluate the efficiency of recognition models, we present the results of Floating Point Operations (FLOPs) and the number of parameters in Table~\ref{tab:eff}. In this study, all trained parameters are accounted for, and the results of FLOPs are computed using a lightweight core library, fvcore\footnote{https://github.com/facebookresearch/fvcore/tree/main\label{fn:flops}}.
The results indicate that although H-CLIP has slightly more parameters than ViFi-CLIP, it has fewer FLOPs. Even the variant based on a multi-layer Transformer has significantly lower FLOPs than the baselines. This demonstrates that H-CLIP not only overcomes the limitations of existing flat learning methods in handling hierarchical video recognition tasks but also represents a cost-effective and efficient approach for video recognition.

\subsection{Visualization}
We randomly sample two video cases from the dataset to intuitively demonstrate the operation process of H-CLIP and the prediction process of hierarchical labels.

\subsubsection{Demonstration of whole process.}
\begin{figure*}[t]
  \centering
   \includegraphics[width=0.98\linewidth]{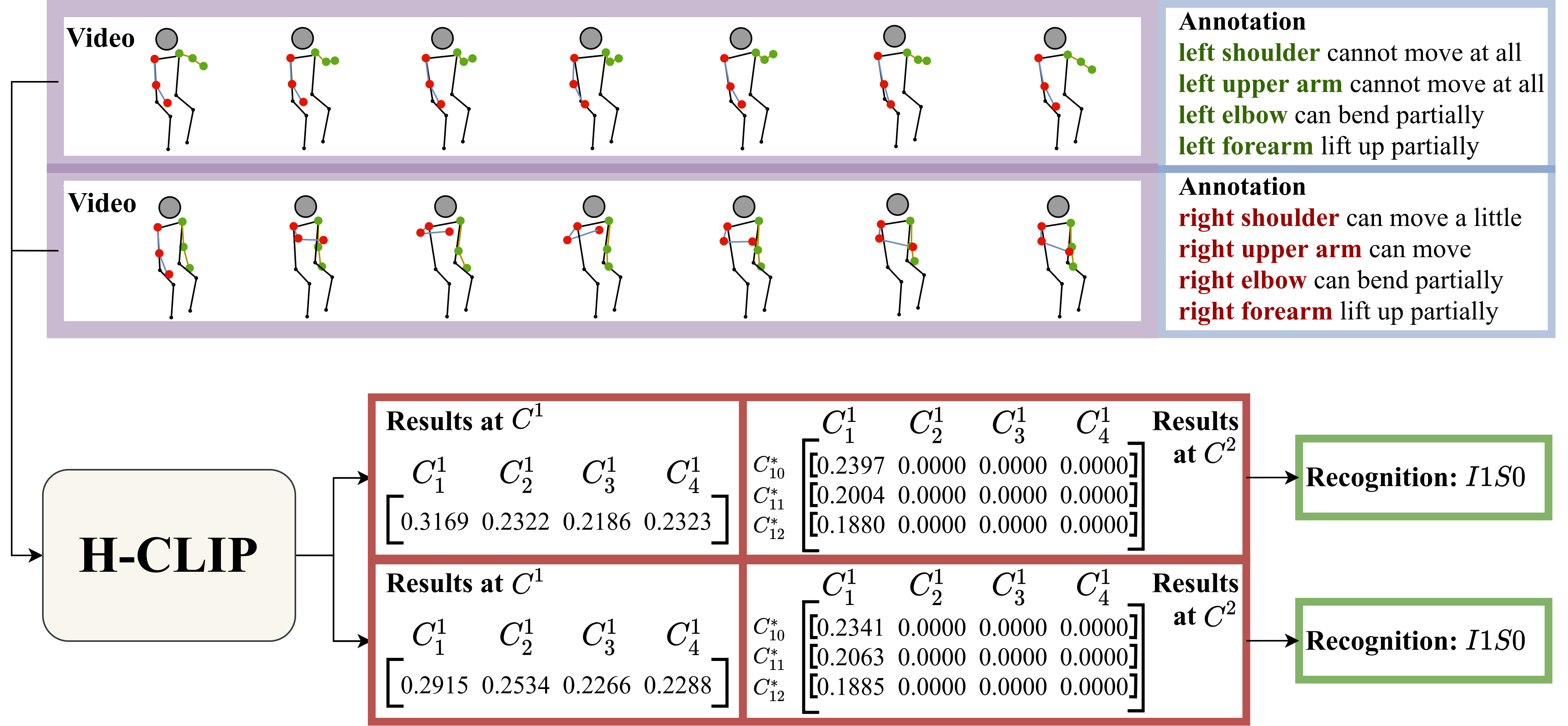}
   \caption{Demonstration of H-CLIP for hierarchical video recognition.}
   \label{fig:demo}
\end{figure*}
Figure \ref{fig:demo} end-to-end hierarchical video recognition process using our proposed H-CLIP model on two example videos from the dataset. 
The dataset contains medical assessment videos where the left and right upper limbs are annotated with text descriptors. These video-text pairs are fed into H-CLIP to produce hierarchical predictions at the coarse $C^1$ level (indicating the medical assessment item category) and fine-grained $C^2$ level (denoting the evaluation degree).
As can be seen, H-CLIP is able to encode both the visual and text modalities to first correctly recognize the assessment item at the high-level $C^1$ category for both videos. This demonstrates the model's ability to capture the overall context and perform coarse-grained classification. More importantly, H-CLIP then further refines the predictions by recognizing the videos into their respective fine-grained $C^2$ labels. This hierarchical multi-level recognition is enabled by interactions across a hierarchy of categories in our proposed H-CLIP framework.

The sequential recognition from coarse to fine-grained levels closely mimics how humans understand layered concepts. By first identifying the general assessment type and then focusing on specific skill detection, the model is able to perform detailed video understanding in a step-wise manner. This example qualitative result highlights the key advantages of our hierarchical approach over flat recognition, showcasing H-CLIP's potential for modeling recursive real-world tasks involving hierarchical reasoning.

\subsubsection{Visualization of hierarchical prediction.}
\begin{figure}[t]
  \centering
   \includegraphics[trim=10 5 5 5, clip=true, width=0.88\linewidth]{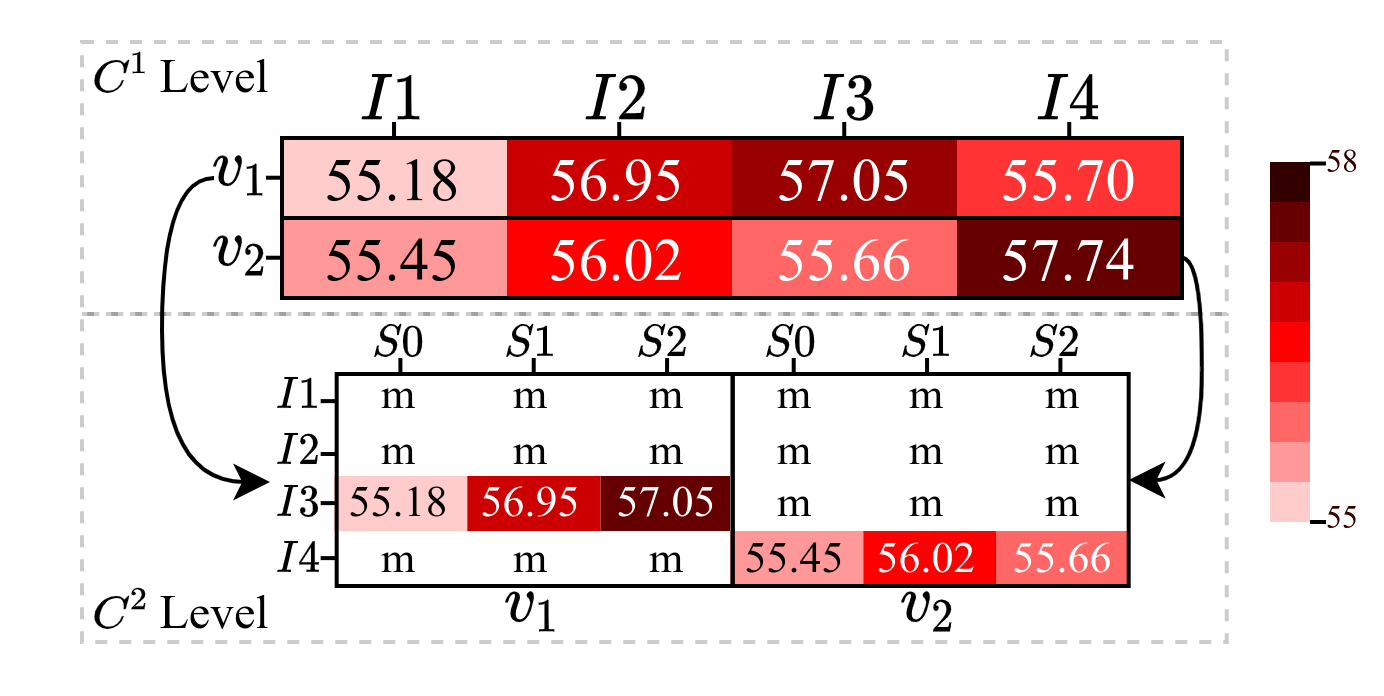}
   \caption{Illustration of hierarchical prediction (\%).}
   \label{fig:case}
\end{figure}

Figure~\ref{fig:case} depicts the detailed process of the hierarchical prediction for the two sampled videos, denoted as $v_1$ and $v_2$. As shown, the ground truth labels are $I3S0$ for $v_1$ and $I4S1$ for $v_2$, indicating their specific medical assessment item and evaluation degree based on the FMA.
It is important to note that for fine-grained medical assessment tasks, the differences between labels at both the $C^1$ and $C^2$ levels can often be minimal and subtle. Small variances in limb positions or motion patterns need to be accurately differentiated. This subtlety underscores the immense challenge of action recognition tasks in medical assessment.
Despite the complexity of the task, our proposed H-CLIP model is able to conduct detailed recognition of the videos' learning objectives. As can be seen, H-CLIP generates predictions that closely match the ground truths, correctly identifying both the general assessment items and evaluation scores. This suggests our model effectively leverages the complementary visual and textual cues through hierarchical multi-modal interactions.

Crucially, H-CLIP also succeeds in the more difficult fine-grained recognition problem at the $C^2$ layer. We believe this is enabled by the hierarchical modeling, which accumulates contextual informationfrom $C^2$ for high-order category learning, and applies a filtering mechanism to constrain the predictions for the specialized $C^2$ recognition. This multi-level reasoning process allows H-CLIP to gradually refine its understanding of the subtle differences between medically similar labels.
In summary, even for this highly intricate hierarchical action recognition problem where distinctions are minute, H-CLIP demonstrates it can still perform detailed learning objective discovery through hierarchical multi-modal modeling and step-wise prediction refinement.

\section{Discussion} 
Unlike mainstream tasks, medical video language understanding demands greater precision, as action differences are often subtle (e.g. $I1S0$ vs $I1S1$, the difference only in that shoulder abduction or elbow flexion occurs "during" or "immediately".). Distinguishing such fine-grained details requires both adequate temporal coverage and precise spatial localization of salient motions. Therefore, a key challenge is accurately predicting assessment items and identifying behavioral differences to determine fine-grained scores given coarse assessments. 
Our work proposes a contrastive video-language approach tailored for this difficult task, demonstrating promising results. However, modeling the intricate complexities of fine-grained medical assessment stretches the capacities of current methods. Significant research is still needed to address the precise spatio-temporal modeling required.
Our method provides a stepping stone towards uncovering discriminative visual cues and learning hierarchical representations for assessment applications. Further innovations in architecture, objectives and incorporation of domain knowledge will be essential to fully tackle this emerging, socially impactful challenge.

\subsection{Limitations}
In current research, hierarchical cases are more prevalent than flat categorization, such as combined action recognition and movie genre classification. When the research object has an inherent hierarchical structure, modeling this taxonomy is often more accurate and reasonable, as evidenced in NLP fields. Beyond medical assessment, video retrieval and recommendation are other promising applications for hierarchical modeling. This allows refinements in search from coarse to fine-grained descriptors, better matching nuanced user needs. Hierarchical modeling also supports generalization, as learning parent-child inheritances can aid in predicting unseen sub-categories given seen parent classes. Furthermore, applying hierarchical modeling to tasks such as temporal localization, video captioning, and video QA can lead to better understanding and reasoning of the visual world, mimicking human-like taxonomic comprehension and ontological reasoning.
However, we must also acknowledge that further progress in hierarchical recognition requires large, diverse datasets with hierarchical labels to verify the value over flat recognition. Annotation cost and complexity pose obstacles. WOverall, our work contributes initial progress in hierarchical video modeling and datasets. Much work remains to develop robust and scalable approaches, but the benefits for visual reasoning and human-like understanding motivate tackling these challenges. We hope our preliminary exploration spurs greater interest in hierarchical video recognition and encourages future data collection, model development, and real-world applications.

\subsection{Future Work}
Hierarchical modeling of visual information aligns with human conceptualization and taxonomy. By incorporating structured relationships, we can strive towards more flexible and human-like understanding of visual information, which is of great importance in the field of computer vision.
With the expansion of intelligent applications, human-centered use cases like healthcare, education, and smart living require higher accuracy. Compared to flat categorization, hierarchical modeling better captures correlations and is more suitable for nuanced real-world needs.
Therefore, developing complete end-to-end applications using hierarchical recognition could provide impactful user services. In healthcare, hierarchical video understanding can enable more professional diagnosis and treatment. Beyond just vision, incorporating diverse modalities like audio and sensor data can provide a fuller analysis of human subjects.
Though real-world deployment necessitates overcoming challenges in robustness, scalability, and acceptability, increased collaboration between domain experts and end-users, iterative system design, objective benchmarking, and qualitative human feedback will facilitate translating these technologies into impactful real-world implementations.

In conclusion, hierarchical modeling offers a promising direction towards flexible, human-like visual intelligence. While progress has been made, much work remains to develop performant and trustworthy applications. Sustained research and open collaboration will be key to unlock the full potential. We hope our preliminary exploration provides a foundation for impactful innovations that benefit society.
\section{Conclusions}

This paper introduces a novel computer vision task, namely hierarchical video recognition, which aims to assign multi-level semantic labels to videos based on a taxonomy. We identify the limitations of existing models that cannot perform simultaneous recognition across hierarchy levels.
To address this, we propose a contrastive video-language approach tailored for hierarchical understanding. Our method leverages inter-level interactions, modeling associations between objects, actions, and hierarchical text descriptors. It further learns dependencies between labels for coherent top-down reasoning.
Moreover, we design a filter mechanism to enforce top-down taxonomic constraints for category predictions based on the affiliations between hierarchical categories. To evaluate our model and catalyze research on hierarchical video recognition, we construct a new real-world dataset of medical assessment videos with over 1,200 examples and fine-grained annotations.
Experiments demonstrate the effectiveness of our approach on this challenging domain, outperforming baselines. Our work provides a strong foundation for advancing hierarchical video modeling.

Looking forward, harnessing hierarchical taxonomic knowledge in combination with multi-modal cues offers an exciting direction for more human-like video understanding. Potential impacts span healthcare, education, robotics, and beyond. To fully unlock these benefits, continued research is needed to address scalability, model interpretability, and real-world variability.
By moving beyond flat classification towards hierarchical recognition, we aim to mimic structured human cognition. Tackling the associated challenges represents the next frontier in video AI. We hope our preliminary exploration and analysis motivates the community to pursue this promising research area and develop innovations that make meaningful societal impacts.








\begin{acks}
The authors would like to express their gratitude to Zhejiang Lab and Shanghai Artificial Intelligence Laboratory for their financial support (K2023KA1BB01, K2022KA1BB01, 2022KA0PI01), which made this research possible. We also extend our sincere thanks to all members of the project team for their invaluable contributions to data collection and their unwavering support throughout the course of this study.
\end{acks}

\bibliographystyle{ACM-Reference-Format}
\bibliography{sample-base}










\end{document}